\documentclass[10pt,twocolumn,letterpaper]{article}

\usepackage{cvpr}      % To produce the REVIEW version
\usepackage{algorithm}
\usepackage{algorithmic}
\usepackage{multirow}
\usepackage{amsthm,amsmath,amssymb}
\usepackage{booktabs}
\usepackage{adjustbox}
\usepackage{amsfonts}
\usepackage{xcolor}

\definecolor{cvprblue}{rgb}{0.21,0.49,0.74}
\usepackage[pagebackref,breaklinks,colorlinks,allcolors=cvprblue]{hyperref}

\def\paperID{13179} % *** Enter the Paper ID here
\def\confName{CVPR}
\def\confYear{2026}

\title{Towards Balanced Multi-Modal Learning in 3D Human Pose Estimation}

\author{
    Mengshi Qi\textsuperscript{1},
    Jiaxuan Peng\textsuperscript{1},
    Xianlin Zhang\textsuperscript{\thanks{Corresponding author: zxlin@bupt.edu.cn.} \hspace{0.5mm} 2, 3}, Huadong Ma\textsuperscript{1} \\
    \small \textsuperscript{1}State Key Laboratory of Networking and Switching Technology, BUPT, China\\
    \small \textsuperscript{2}School of Digital Media \& Design Art, BUPT, China\\
    \small \textsuperscript{3}Beijing Key Laboratory of Intelligent Computing and Experience Innovation for Sci-Fi Visual Spaces\\
    \small \texttt{\{qms, pjx, zxlin, mhd\}@bupt.edu.cn}
}

\begin{document}
\maketitle
\begin{abstract}
3D human pose estimation (3D HPE) has emerged as a prominent research topic, particularly in the realm of RGB-based methods. However, the use of RGB images is often limited by issues such as occlusion and privacy constraints. Consequently, multi-modal sensing, which leverages non-intrusive sensors, is gaining increasing attention. Nevertheless, multi-modal 3D HPE still faces challenges, including modality imbalance. In this work, we introduce a novel balanced multi-modal learning method for 3D HPE, which harnesses the power of RGB, LiDAR, mmWave, and WiFi. Specifically, we propose a Shapley value-based contribution algorithm to assess the contribution of each modality and detect modality imbalance. To address this imbalance, we design a modality learning regulation strategy that decelerates the learning process during the early stages of training. We conduct extensive experiments on the widely adopted multi-modal dataset, \emph{MM-Fi}, demonstrating the superiority of our approach in enhancing 3D pose estimation under complex conditions. Our source code is available at \href{https://github.com/MICLAB-BUPT/AWC}{https://github.com/MICLAB-BUPT/AWC}.
\end{abstract}    
\section{Introduction}
\begin{figure}[!t]
    \centering
    \includegraphics[width=0.9\linewidth]{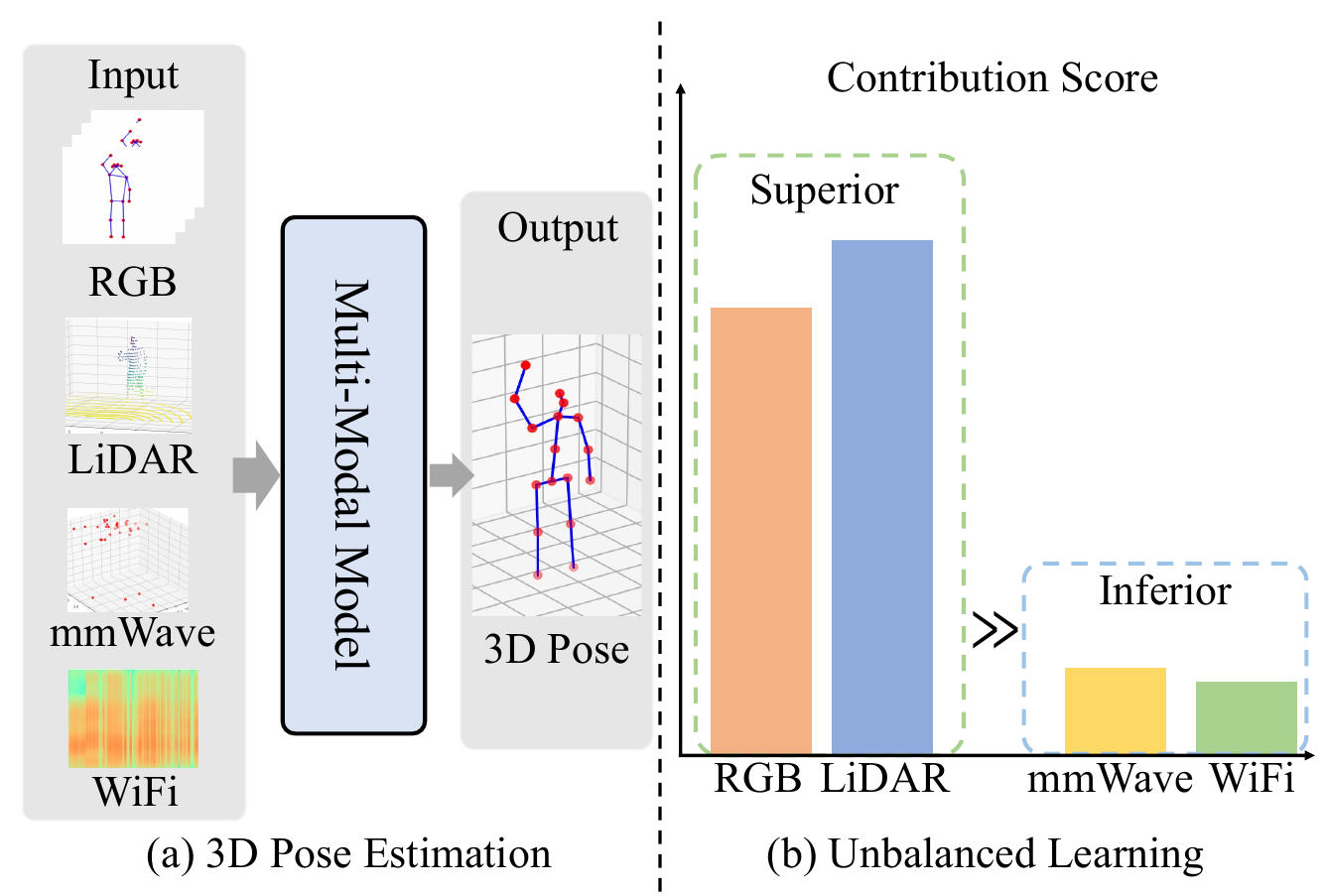}
    \caption{(a) Multi-modal 3D HPE. (b) In end-to-end training, modality imbalance arises where dominant modalities with higher scores suppress the training of others in MM-Fi~\cite{yang2024mm}.}
    \label{fig1}
    \vspace{-4mm}
\end{figure}

3D human pose estimation (3D HPE) recovers 3D coordinates of human joints from various input sources. It has gained significant research attention due to its applications in human-robot interaction~\cite{garcia2019human, zimmermann20183d}, action assessment~\cite{qi2025action} and computer animation~\cite{mehta2017vnect}. Specifically, in the rehabilitation context, where a variety of sensors and monitoring devices are deployed in the surroundings to detect patient action, 3D HPE models can act as an indispensable tool for supervising and verifying the correctness of patients' exercises, ensuring adherence to established standards. 
Existing methods primarily focus on camera-based inputs (\textit{i.e.,} RGB images and videos) due to their accessibility and abundant human body information. However, camera-based approaches encounter limitations under occlusion scenarios and necessitate complex spatial conversion from 2D to 3D, dependent on accurate camera parameters. Hence multi-modal human sensing emerges as a promising approach for addressing complex scenarios by leveraging diverse sensor modalities. Wearable sensors are constrained by user compliance, hindering their practical adoption in everyday scenarios. Therefore, non-intrusive sensors such as LiDAR, mmWave radar, and WiFi offer advantages in terms of illumination invariance and user convenience, as shown in Figure~\ref{fig1} (a). By fusing information from RGB and non-intrusive sensors, we can enhance downstream task performance by exploiting both complementary and redundant information~\cite{wang2015mmss}. Nevertheless, existing methods~\cite{chen2023immfusion, an2022fast, he2024video, zheng2022multi, furst2021hperl} primarily rely on one or two modalities, leaving three or more modalities as an unexplored area.

Although multi-modal models are designed for extracting information from all modalities, dominant modalities suppress the optimization of less dominant modalities in multi-modal learning as noted in~\cite{wang2020makes, peng2022balanced}, leading to modality imbalance, shown in Figure~\ref{fig1} (b). While modality modulation techniques~\cite{peng2022balanced, li2023boosting} mitigate this issue by interfering with the learning of the dominant, they can degrade overall performance in certain scenarios, as discussed in \cite{wei2024diagnosing}. This is attributed to their neglect of the intrinsic limitations of different modalities' information capacity, resulting in a failure to achieve an effective balance. Consequently, the challenge lies in balancing modality optimizations without compromising the learning of dominant modalities. 

Furthermore, existing balancing methods~\cite{wu2022characterizing, peng2022balanced, li2023boosting, ReconBoost, fan2023pmr} are specifically tailored for discriminative tasks that rely on cross-entropy loss or involve explicit class attribution, making them inherently unsuitable for regression tasks. In addition, several approaches~\cite{wei2024innocent, guo2024classifier, wang2020makes} introduce auxiliary uni-modal components, such as uni-modal heads, to facilitate optimization. However, this design choice typically results in increased model complexity and additional optimization for those auxiliary parameters.

To address the aforementioned limitations, we propose a novel multi-modal learning framework that achieves balanced optimization across all modalities in 3D HPE without introducing additional model complexity. Our approach is built upon two key components: a Shapley value-based contribution assessment and a Fisher Information Matrix (FIM)-guided adaptive weight regularization.

First, we employ a Shapley value-based algorithm to quantify each modality’s contribution during training. 
Unlike previous balancing techniques that rely on auxiliary uni-modal heads or additional network branches, the Shapley analysis~\cite{shapley1953value} is fully parameter-free and can be directly computed within existing multi-modal fusion architectures. 
This enables the model to dynamically monitor the relative importance of modalities, serving as a reliable indicator of their optimization status and guiding adaptive balancing throughout training.

Second, we introduce a weight constraint loss that leverages the Fisher Information Matrix (FIM)~\cite{fisher1925theory} to adaptively regulate parameter updates based on their empirical importance. The FIM provides a principled, data-driven estimate of parameter sensitivity by approximating the expected squared gradient magnitude, which is an indicator of how much each parameter contributes to loss reduction. Crucially, in multimodal settings, dominant modalities tend to produce larger gradients in the beginning of training~\cite{huang2025adaptive, achille2018critical,huang2023fuller} and thus higher FIM values, while weaker modalities yield smaller ones. By penalizing deviations from initial parameters in proportion to their FIM-weighted importance, our loss applies stronger regularization to dominant modalities and milder constraints to inferior ones. This adaptive mechanism mitigates optimization imbalance: it curbs overly aggressive updates from strong modalities without excessively suppressing weaker ones, thereby fostering more balanced and robust multi-modal learning.

% \pjx{To address the issue of modality imbalance in regression tasks and overcome the aforementioned limitations, we propose a novel multi-modal learning framework that achieves balanced optimization across all modalities in 3D HPE. Specifically, we utilize a Shapley value-based contribution assessment algorithm to quantify the contribution of each modality during training. This method is compatible with diverse and complex multi-modal fusion architectures, requires no auxiliary parameters, and incurs negligible computational overhead. The computed contribution scores serve as indicators of each modality’s current optimization status, enabling adaptive control the balancing process. Moreover, we introduce a regularization term, referred to as the weight constraint loss, which modulates the learning process by penalizing parameter updates in proportion to their estimated importance. Specifically, the importance of each parameter is evaluated using the Fisher information matrix~\cite{fisher1925theory}, and parameters with higher importance, which are primarily associated with dominant modalities~\cite{huang2025adaptive}, are subjected to stronger constraints during optimization. Consequently, this mechanism effectively slows down the learning of dominant modalities, while reducing the risk of overfitting to noisy or less informative updates from underrepresented modalities.}

Our main contributions are summarized as follows:
\par\textbf{(1)} We propose a novel balanced multi-modal model for 3D human pose estimation that seamlessly integrates RGB, LiDAR, mmWave, and WiFi modalities.
\par\textbf{(2)} We employ a Shapley value-based contribution assessment algorithm, which leverage Shapley Value and Pearson correlation coefficient to quantify each modality’s utility, providing an effective indicator of the model’s training imbalance.
\par\textbf{(3)} We introduce a novel weight regularization term, adaptive weight constraint (AWC) loss, to dynamically balance the optimization of different modalities during training, without introducing auxiliary parameters.
\par\textbf{(4)} We conduct extensive experiments on the largest multi-modal 3D HPE dataset, \emph{MM-Fi}~\cite{yang2024mm}, demonstrating the superiority of our approach over baseline methods.

\begin{figure*}
    \centering
    \includegraphics[width=\linewidth]{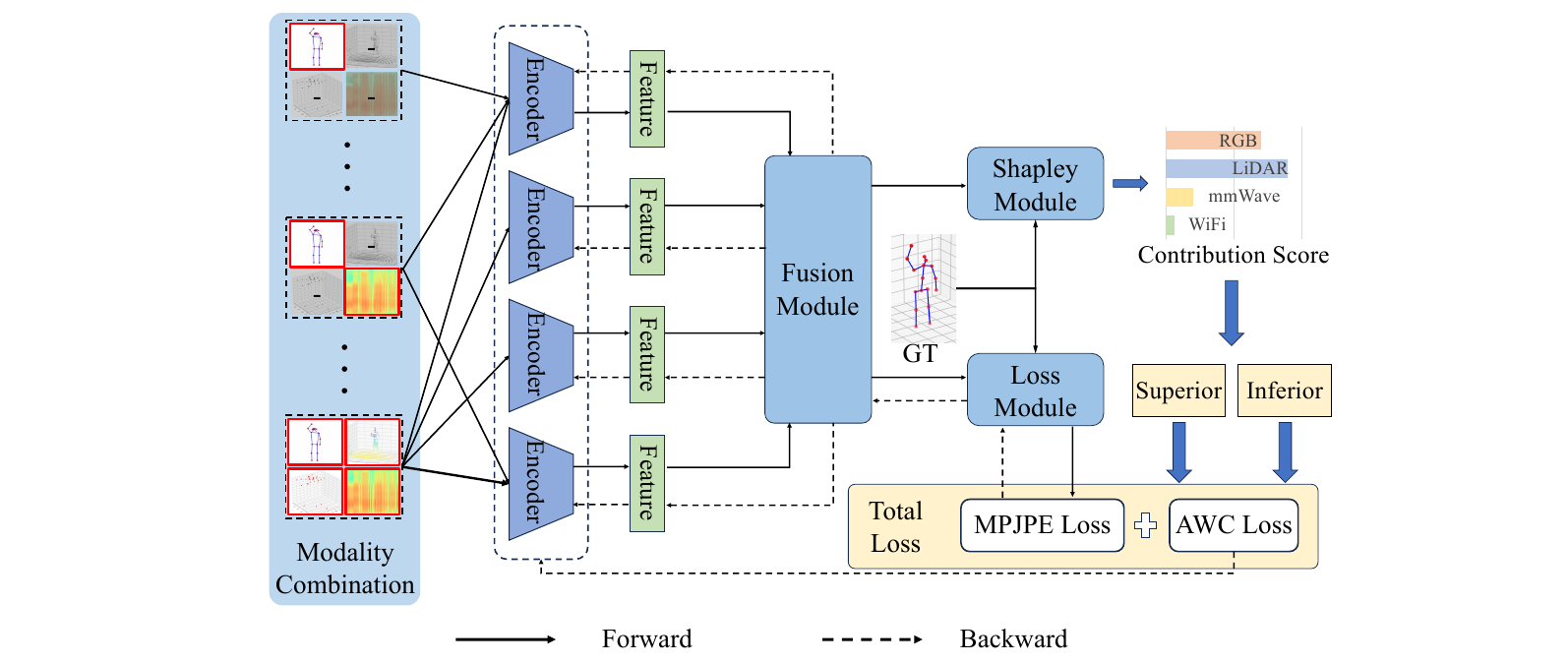}
    \caption{Illustration of our proposed methods. Modality contributions are assessed via Shapley Module, and an adaptive weight constraint (AWC) loss, weighted by Fisher information, regularizes encoder parameter updates to slow dominant modalities and protect inferior ones during the critical learning window.}
    \label{model}
    \vspace{-4mm}
\end{figure*}
\section{Related Work}

\noindent{\bf 3D Human Pose Estimation.}
Video data plays a central role in many computer vision tasks~\cite{qi2019sports,qi2020imitative,qi2020stc,qi2019stagnet,qi2019attentive,qi2019ke} and has become a widely used modality for 3D HPE. 
Existing approaches can be broadly categorized into two classes: directly estimating 3D joints~\cite{tekin2016direct, pavlakos2017coarse, wehrbein2021probabilistic, moon2019camera}, and 2D-to-3D lifting~\cite{cai2019exploiting, peng2024ktpformer, li2024hourglass, xu2024finepose}.
% Video data plays a central role in many computer vision tasks~\cite{qi2019sports,qi2020imitative,qi2020stc,qi2019stagnet,qi2019attentive,qi2019ke} and and has also become the predominant modality for 3D HPE, which can be categorized into two classes: directly estimating 3D joints \cite{tekin2016direct, pavlakos2017coarse, wehrbein2021probabilistic, moon2019camera} and 2D-to-3D lifting \cite{cai2019exploiting, peng2024ktpformer, li2024hourglass, xu2024finepose}.
% Currently, the predominant focus of 3D HPE is on vision-based methods, categorized into two classes: directly estimating 3D joints \cite{tekin2016direct, pavlakos2017coarse, wehrbein2021probabilistic, moon2019camera} and 2D-to-3D lifting \cite{cai2019exploiting, peng2024ktpformer, li2024hourglass, xu2024finepose}.
Recently, LiDAR has been applied to 3D HPE due to its robustness, which is utilized in autonomous driving~\cite{zheng2022multi, cong2022stcrowd}. MmWave-based HPE has gained increasing attention in research community~\cite{zhu2024probradarm3f}, driven by the demand for privacy-preserving technologies. WiFi-based sensing has also emerged as a promising area of research~\cite{arshad2017wi, zou2018robust, geng2022densepose} due to its popularity. Although RGB-based methods are the majority in research community, multi-modal methods could sacrifice the computational complexity for better performance and robustness. In this work, we aim to integrate these modalities to enhance 3D HPE performance.

\noindent{\bf Multi-Modal Learning.}
Multi-modal learning~\cite{qi2025dc,qi2025robust,Lv_2026_CVPR,qi2021semantics,qi2017online} has attracted significant attention across diverse perception and understanding tasks~\cite{lv2024sgformer,lv2025t2sg,deng2025global,wang2025pitn,ye2025safedriverag,qi2025synergistic,qi2021latent}.
However, training such models remains challenging due to complex interactions among modalities.
Several works~\cite{wang2020makes, peng2022balanced, guo2024classifier} focus on end-to-end training, revealing issues like modality imbalance or competition, causing the suboptimal performance~\cite{peng2022balanced}. To address this issue, several methods~\cite{li2023boosting, peng2022balanced} are proposed to quantify modality imbalance and improve training by modulating the gradients of dominant modalities. PMR~\cite{fan2023pmr} introduces a prototype learning approach based on class-specific representations in classification tasks. MMPareto~\cite{wei2024innocent} incorporates the concept of the Pareto front into multi-modal learning and optimizes the multi-modal gradient based on joint uni-modal gradients. Although existing methods have made significant progress in mitigating such imbalance, most are tailored to classification tasks and often rely on additional learnable parameters. In contrast, our work addresses modality imbalance in the context of regression, specifically 3D human pose estimation, by leveraging Shapley value-based contribution assessment and an adaptive regularization to enable balanced optimization. Crucially, our approach introduces no extra learnable parameters.

%  As a result, the problem of modality imbalance in regression tasks remains largely underexplored. Recently, CGGM~\cite{guo2024classifier} proposed a classifier-guided gradient modulation strategy applicable to regression tasks. This method introduces auxiliary uni-modal decoders and heads to modulate the directions and magnitude of multi-modal gradients, but at the cost of additional parameters and optimization for those modules. In contrast, 
\section{Proposed Approach}

% As illustrated in Figure~\ref{model}, balanced multi-modal learning consists of two components. First, a Shapley value-based contribution algorithm calculates the uni-modal contribution scores based on Pearson correlation and Shapley value. Second, an adaptive regularization term dynamically balances the optimization of different modalities.
\subsection{Overview}
\noindent\textbf{Problem Definition.} In the multi-modal 3D human pose estimation problem, given the four modality types $\mathcal{M} := \{RGB \,(R), LiDAR \,(L), mmWave\,(M), WiFi \,(W)\}$, as depicted in Figure~\ref{fig1}(a), our objective is to estimate the corresponding 3D coordinates of the $j$-th human joints, denoted as $\hat{y} = {\rm MM}(\mathcal{M})$, $\hat{y} \in \mathbb{R}^{j \times 3}$, where ${\rm MM}(\cdot)$ represents the multi-modal model. Specifically, the RGB inputs $X_{R} = \{p^{2d}_i\}_{i=0}^{N}$, $p^{2d}_i \in \mathbb{R}^{j \times 2}$, consist of $N$ frames from a video, each containing $j$ 2D human joints extracted from RGB images. The LiDAR point cloud is represented as $X_{L} = \{p_i\}_{i=0}^{N_{L}}$, $p_i \in \mathbb{R}^{N_{L} \times 3}$, where $N_{L}$ is the total number of LiDAR points in a single frame. The mmWave radar point cloud is represented as $X_{M} = \{p_i\}_{i=0}^{N_{M}}$, $p_i \in \mathbb{R}^{N_{M} \times d}$, where each point $p_i = (x, y, z, D, I)$ includes 3D coordinates, Doppler velocity $D$, and signal intensity $I$, with $N_{M}$ denoting the number of mmWave points. WiFi CSI data is denoted as $X_{W} = s$, $s \in \mathbb{R}^{a \times c \times t}$, where $a$ represents the number of WiFi antennas, $c$ denotes the number of subcarriers per antenna, and $t$ refers to the sampling frequency.

Our proposed method, as illustrated in Figure~\ref{model}, comprises two components. Shapley value-based contribution algorithm assesses the modalities by computing uni-modal contribution scores and the adaptive weight constraint loss balances the optimization of modalities. Initially, our model employs modality-specific encoders to extract features from the respective data sources. Subsequently, a multi-modal fusion module combines the features from these modality-specific branches, and a pose regression head is finally applied to predict the final results.

\subsection{Shapley Value-Based Contribution Algorithm}
Shapley Value~\cite{shapley1953value} was introduced in coalition game theory to address profit distribution by calculating each player's marginal contribution to the group, ensuring fair distribution. Prior works on modality imbalance~\cite{hu2022shape, li2023boosting, wei2024enhancing} have introduced the Shapley value into multimodal learning to assess modality contributions in classification tasks but limited in the regression. Specifically, the contribution score $\phi^m$ for modality $m$ is obtained by calculating the profits provided by all permutations of $\mathcal{M}\setminus\{m\}$ cooperated with $m$. Let $S$ denotes a subset of $\mathcal{M}\setminus\{m\}$, the calculation is defined as:
\begin{equation}\label{eq2}
    \phi^m(\mathcal{M})=\sum_{S\subseteq \mathcal{M}\setminus\{m\}}\frac{|S|!(|\mathcal{M}|-|S|-1)!}{|\mathcal{M}|!}V(S, m),
\end{equation}
where $V(S, m) = s(y, {\rm MM}(S\cup\{m\}))-s(y, {\rm MM}(S))$. $V(S, m)$ quantifies the additional profit gained by incorporating modality $m$ into subset $S$, and $s(\cdot,\cdot)$ is leveraged to calculate the Shapley value profit by the ground truth $y$ and the prediction $\hat{y}={\rm MM}(\mathcal{M})$. 

\noindent\textbf{Shapley Value in classification tasks.} In classification settings, the cross-entropy loss is commonly used as the profit function $s(\cdot,\cdot)$ in prior methods~\cite{hu2022shape, li2023boosting, wei2024enhancing}. However, in regression tasks, directly applying the MSE or MAE loss between ground truths $y$ and predictions $\hat{y}$ as the profit function becomes problematic. 

For illustrative purposes, we consider a standard fusion strategy, feature concatenation. The input samples $x^R$, $x^L$, $x^M$, and $x^W$ are first passed through modality-specific encoders $\varphi^R$, $\varphi^L$, $\varphi^M$, and $\varphi^W$, yielding feature representations $f^R$, $f^L$, $f^M$, and $f^W$, where $R$, $L$, $M$, and $W$ are the RGB, LiDAR, mmWave, and WiFi modalities, respectively. These features are then concatenated along the feature dimension and fed into a linear output layer. The prediction is formulated as:
\begin{equation}\label{concat}
    \hat{y} = w\left[f^R; f^L; f^M; f^W\right] + b,
\end{equation}
where $w\in\mathbb{R}^{d_{out}\times(d_{\varphi_R} + d_{\varphi_L} + d_{\varphi_M} + d_{\varphi_W})}$ and $b$ represent the weight and bias parameters of the linear layer, respectively. $\hat{y}$ denotes the task-specific output, and $d_{out}$ corresponds to the number of classes in classification or the number of joint coordinates in 3D human pose estimation. By decomposing the weight matrix $w$ into $\left[w^R;w^L;w^M;w^W\right]$, Equation~\ref{concat} can be reformulated as:
\begin{equation}\label{pred_res}
\begin{aligned}
    \hat{y} &= w^R f^R + w^L f^L + w^M f^M + w^W f^W + b \\
    % \hat{y} &= w^R \cdot f^R + w^L \cdot f^L + w^M \cdot f^M + w^W \cdot f^W + b \\
    & = \hat{y}^R + \hat{y}^L + \hat{y}^M + \hat{y}^W.
\end{aligned}
\end{equation}

The final prediction of the model, whether for classification or regression tasks, is derived as the sum of individual predictions from the four modalities, as formulated in Equation~\ref{pred_res}. However, the way each modality contributes to the overall performance differs significantly between the two types of tasks. In classification tasks, if a certain modality provides weak or noisy features and thus yields poor discriminative power, its predicted logits tend to be close to a uniform distribution. As a result, the absence or inclusion of this modality when assessing its marginal contribution to the subset $S$ has minimal impact on the softmax-normalized probabilities, since adding or removing a value approximate to a constant term from the logits does not alter the normalized probability distribution. This property allows classification tasks to adopt the cross-entropy loss as the profit function $s(\cdot, \cdot)$. 

\begin{figure}
    \centering
    \includegraphics[width=\linewidth]{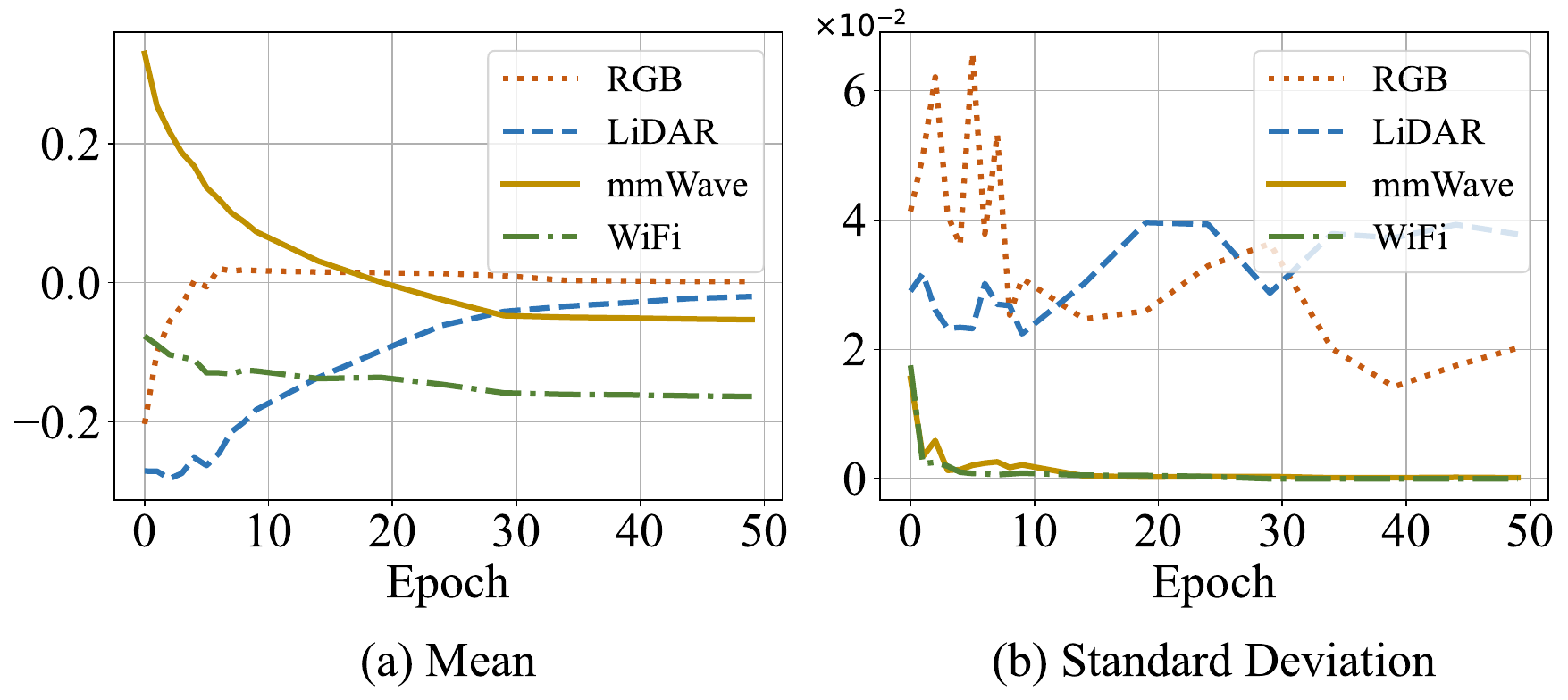}
    \caption{(a) Mean and (b) standard deviation of human joint coordinate predictions sampled from MM-Fi~\cite{yang2024mm} during the training process. Modalities like mmWave and WiFi show near-zero standard deviations, indicating their predictions collapse to nearly constant values and contribute little to pose estimation.}
    \label{mean_std}
    \vspace{-3mm}
\end{figure}

\noindent\textbf{Shapley Value in regression tasks.} In contrast, regression tasks exhibit a fundamentally different behavior. When using metrics such as MSE or MAE to evaluate prediction accuracy, the model aims to minimize the numerical discrepancy between $\hat{y}$ and $y$. In this context, directly using MSE as the profit function $s(\cdot,\cdot)$ introduces a bias toward modalities with larger output magnitudes, as their contribution to the overall error becomes disproportionately large. Specifically, modalities with weaker features tend to produce more stable and less variable predictions as shown in Figure~\ref{mean_std}, which can be mistakenly interpreted as having higher reliability or informativeness when leveraging Euclidean distance to evaluate the contribution. Consequently, using MSE or MAE as the profit function for computing Shapley profit in regression tasks may not accurately reflect the true informative value of each modality. This observation highlights the need for alternative evaluation metrics when analyzing modality contributions in regression-based multi-modal learning systems.

Based on the above analysis and observations in Figure~\ref{mean_std}, we calculate the Pearson correlation coefficient of each joint coordinate value between $y$ and $\hat{y}$ along the batch dimension, which serves as $s(\cdot, \cdot)$  and is formulated as:
\begin{equation}\label{eq1}
    s(y, \hat{y}) = \sum_{i=1}^{j\times3} \rho(y_i, \hat{y}_i),
\end{equation}
where $j$ represents the number of human joints and $\rho(y_i, \hat{y}_i)$ is the Pearson correlation, formulated as:
\begin{equation}\label{pear}
    \rho(y_i, \hat{y}_i) = \frac{cov(y_i, \hat{y}_i)}{\sigma_{y_i}\cdot\sigma_{\hat{y}_i}},
\end{equation}
where $cov(y_i, \hat{y}_i)$ is the covariance of $y_i$ and $\hat{y}_i$, and $\sigma_{y_i}$ and $\sigma_{y_i}$ are the standard deviations of $y_i$ and $\hat{y}_i$, respectively. When modalities are absent from $S$, we apply zero-padding to their features, ensuring network compatibility. This process iteratively computes the contribution of each modality across all potential combinations, culminating in all contribution scores for all modalities. By employing Pearson correlation as a substitute for logits in classification models, we successfully extend uni-modal contribution analysis to regression tasks.

% \begin{algorithm}[tb]
% \caption{Balanced Multi-Modal Learning}
% \label{alg}
% \textbf{Input}: Dataset $\mathcal{D} = \{x_i^m, y_i\}_{i=1,2...n}, m\in\mathcal{M}$, the number of epochs $\mathcal{N}$, the number of batches $\mathcal{B}$, learning window $K$ \\
% \textbf{Parameter}: multi-modal parameters $\theta$

% \begin{algorithmic} %[1] enables line numbers
% \FOR{$n = 0, 1, \ldots, \mathcal{N}-1$}
% \IF{$n<K$}
% \STATE Calculate FIM $[\mathcal{I}]$ using Equation~\ref{eq:fim};
% \ENDIF
% \FOR{$m = 0, 1, \ldots, \mathcal{B}-1$}
% \STATE Sample a batch $b$ in $\mathcal{D}$;
% \STATE Forward using $\hat{y} = {\rm MM}(b)$;
% \STATE Calculate the contribution scores using Equation~\ref{eq2};
% \IF{$n<K$}
%     \STATE Calculate the final loss $\mathcal{L}_{total}= \mathcal{L}+\mathcal{L}_{AWC}$;
% \ELSE
%     \STATE Calculate the final loss $\mathcal{L}_{total} = \mathcal{L}$;
% \ENDIF
% \STATE Backward with $\mathcal{L}_{total}$ and update the parameters.
% \ENDFOR
% \ENDFOR
% \end{algorithmic}
% \end{algorithm}

\subsection{Adaptive Weight Constraint Regularization}
% \pjx{Previous studies~\cite{huang2025adaptive, achille2018critical} identify a critical learning window in the early training, during which the model acquires the majority of its task-relevant information. Within this window, dominant modalities typically learn rapidly, driving a sharp decline in the overall task loss. However, this fast convergence disproportionately suppresses the gradient magnitudes of inferior modalities. As a result, these modalities often converge to a suboptimal state. To mitigate this imbalance, it is crucial to moderate the learning dynamics of dominant modalities during the critical window, thereby creating sufficient opportunity for inferior modalities to learn and achieve more balanced multi-modal learning.}

Existing methods~\cite{wei2024innocent, huang2025adaptive} typically modulate either the direction or the magnitude of gradients in isolation, despite the fact that both jointly govern how modality-specific encoders are updated during training. Moreover, these approaches often focus exclusively on information-sufficient (i.e., dominant) modalities and overlook the risk of overfitting to noisy signals in weaker ones.

To address these limitations, we propose an \textit{adaptive weight constraint} (AWC) regularization method, which simultaneously regulates both the direction and magnitude of parameter updates for all modalities. Our key idea is to differentiate regularization strengths based on each modality’s contribution significance, taking into account their distinct training dynamics. Specifically, we firstly partition the full modality set $\mathcal{M}$ into two disjoint subsets: one for superior modalities and one for inferior modalities. This partitioning is achieved by applying K-Means clustering to the Shapley scores of the four modalities. The cluster with the higher mean Shapley score is designated as the set of \textbf{superior} modalities, denoted $\mathcal{M}_{\mathcal{S}}$, and is assigned a stronger regularization coefficient $\alpha_{\mathcal{S}}$. Conversely, the lower-scoring cluster forms the set of \textbf{inferior} modalities, $\mathcal{M}_{\mathcal{I}}$, regularized with $\alpha_{\mathcal{I}}$. This adaptive strategy ensures that both strong and weak modalities are appropriately constrained according to their reliability, thereby mitigating overfitting while preserving informative updates.

\noindent\textbf{Adaptive weight constraint loss. } An adaptive weight constraint loss term, denoted as $\mathcal{L}_{\text{AWC}}$, is introduced to regularize the deviation of current parameters from their initial values, weighted by their importance as measured by the diagonal of the Fisher Information Matrix (FIM). Formally, it is defined as:
\begin{equation}
    \label{awc_loss}
    \mathcal{L}_{\text{AWC}} = 
    \sum_{m\in\mathcal{M}}
    \left[
    \alpha_{\mathcal{S}} \cdot \mathbf{1}_{\{m\in\mathcal{M}_{\mathcal{S}}\}} + 
    \alpha_{\mathcal{I}} \cdot \mathbf{1}_{\{m\in\mathcal{M}_{\mathcal{I}}\}}
    \right]
    \cdot \mathcal{L}_{W}^{m},
\end{equation}
where
\begin{equation}
    \mathcal{L}_{W}^{m} = \sum_i \frac{[\mathcal{I}_{\mathcal{D}}]_{ii} \, (\theta_{t,i}^m - \theta_{0,i}^{m,*})^2}{2}.
\end{equation}
Here, $\theta_{0,i}^{m,*}$ denotes the $i$-th parameter of modality $m$ at the beginning of the current epoch (i.e., step 0), and $\theta_{t,i}^m$ is its value at update step $t$. The term $\mathcal{L}_{W}^{m}$ represents the modality-specific encoder regularization loss, designed to constrain the magnitude and the direction of parameter updates for modality $m$. The term $[\mathcal{I}_{\mathcal{D}}]_{ii}$ represents the diagonal approximation of the Fisher Information Matrix computed on dataset $\mathcal{D}$:
\begin{equation}\label{eq:fim}
    [\mathcal{I}]_{ii} = \frac{1}{|\mathcal{D}|} \sum_{(x_n, y_n)\in\mathcal{D}} \left( \frac{\partial \mathcal{L}(x_n, y_n; \theta_0^*)}{\partial \theta_i} \right)^2,
\end{equation}
which is evaluated using the initial parameters $\theta_0^*$ at the beginning of each epoch.

% \pjx{
% Although $\mathcal{L}_{\text{AWC}}$ is applied to all modalities, its effective strength is adaptively determined by the data-driven Fisher Information Matrix (FIM). 
% Inferior modalities, with smaller gradients and lower diagonal FIM entries, receive weaker regularization, while superior ones with larger gradients are more strongly constrained. This adaptive weighting naturally balances optimization, preventing dominant modalities from overwhelming training while allowing weaker ones to learn informative signals without overfitting to noise.
% }

\noindent\textbf{Learning window. } Furthermore, following prior work~\cite{huang2025adaptive, achille2018critical} showing that most task-relevant information is acquired during early training, we apply $\mathcal{L}_{\text{AWC}}$ only within this initial phase. 
In practice, the first $K$ epochs are defined as the learning window, which is a tunable hyperparameter. Restricting $\mathcal{L}_{\text{AWC}}$ to this window mitigates the rapid dominance of strong modalities in early optimization, giving weaker modalities sufficient opportunity to acquire useful representations and achieve balanced multimodal learning.

\begin{table*}[!t]\small
\centering
% \footnotesize
\caption{Comparisons of our proposed method and existing balancing multi-modal learning methods on MM-Fi. - denotes the results are inapplicable. The lower is better; the best results are highlighted in \textbf{bold}.
}
\begin{tabular}{lcccccc}
\toprule
&
  \multicolumn{2}{c}{Protocol 1} &
  \multicolumn{2}{c}{Protocol 2} &
  \multicolumn{2}{c}{Protocol 3} \\ 
  \cmidrule(lr){2-3}  \cmidrule(lr){4-5} \cmidrule(lr){6-7}
  \multirow{-2.5}{*}{Methods}      & MPJPE    & PA-MPJPE   & MPJPE  & PA-MPJPE  & MPJPE   & PA-MPJPE \\   
    \cmidrule{1-7}
MM-Fi~\cite{yang2024mm} & 72.90 & 47.70 & 69.50 & 43.10 & 89.80 & 63.20 \\
X-Fi~\cite{chen2025xfi} & - & - & - & - & 88.60 & 57.10 \\
HPE-Li~\cite{d2024hpe} & 152.71 & 94.39 & 162.42 & 89.18 & 149.42 & 92.52 \\
    \midrule
Concatenation & 53.87 & 35.09 & 52.08 & 30.97 & 48.17 & 32.18 \\
 $\ +$ G-Blending \cite{wang2020makes} & 58.40 & 37.20 & 53.48 & 34.30 & 53.13 & 33.28 \\
 $\ +$ OGM-GE \cite{peng2022balanced} & 55.51 & 35.92 & 52.37 & 32.58 & 51.68 & 32.84 \\
 $\ +$ AGM \cite{li2023boosting} & 55.80 & 38.10 & 54.21 & 37.95 & 53.88 & 36.30 \\
 $\ +$ Modality-level \cite{wei2024enhancing} & 53.24 & 34.81 & 55.31 & 30.88 & 53.98 & 31.85 \\
 \cmidrule(lr){2-3} \cmidrule(lr){4-5} \cmidrule(lr){6-7}
$\ +$ \textbf{Ours} & \textbf{51.16} & \textbf{34.46} & \textbf{50.71} & \textbf{30.85} &  \textbf{47.55} & \textbf{31.79}  \\
\midrule
Attention & 53.35 & 35.20 & 51.01 & 31.15 & 49.97 & 32.33 \\
 $\ +$ G-Blending \cite{wang2020makes} & 57.14 & 38.98 & 53.85 & 34.52 & 53.93 & 34.36 \\
 $\ +$ OGM-GE \cite{peng2022balanced} & - & - & - & - & - & - \\
 $\ +$ AGM \cite{li2023boosting} & 58.40 & 39.91 & 61.28 & 41.24 & 53.60 & 36.86 \\
 $\ +$ Modality-level \cite{wei2024enhancing} & 53.33 & 35.23 & 50.48 & 31.47 & 50.71 & 33.46 \\
 \cmidrule(lr){2-3} \cmidrule(lr){4-5} \cmidrule(lr){6-7}
$\ +$ \textbf{Ours} & \textbf{51.29}  & \textbf{34.65} & \textbf{50.42} & \textbf{30.85} &  \textbf{49.08} & \textbf{32.10}  \\
\bottomrule
\end{tabular}
\label{table1}
\vspace{-3mm}
\end{table*}

\begin{algorithm}[tb]
\caption{Balanced Multi-Modal Learning}
\label{alg}
\textbf{Input}: 
Dataset $\mathcal{D} = \{(x_i^m, y_i)\}_{i=1}^N$, where $m \in \mathcal{M}$;  
Number of epochs $\mathcal{N}$;  
Number of batches per epoch $\mathcal{B}$;  
Learning window length $K$ (in epochs). \\
\textbf{Initialize}: Multi-modal model parameters $\theta$.

\begin{algorithmic} %[1] enables line numbers
\FOR{epoch $e = 0$ \textbf{to} $\mathcal{N}-1$}
\IF{$e < K$}
\STATE Compute FIM $[\mathcal{I}]$ using Eq.~\ref{eq:fim}.
\ENDIF
\FOR{batch index $b = 0$ \textbf{to} $\mathcal{B}-1$}
\STATE Sample a mini-batch $\mathcal{D}_b \subset \mathcal{D}$.
\STATE Forward pass: $\hat{y} \gets \mathrm{MM}(\mathcal{D}_b; \theta)$.
\STATE Calculate the contribution scores using Eq.~\ref{eq2}.
\STATE Compute task loss $\mathcal{L}_\text{MPJPE}$.
\IF{$e < K$}
\STATE Compute AWC loss $\mathcal{L}_{\text{AWC}}$ as in Eq.~\ref{awc_loss}.
\STATE Set total loss: $\mathcal{L}_{\text{total}} \gets \mathcal{L}_\text{MPJPE} + \mathcal{L}_{\text{AWC}}$.
\ELSE
\STATE Set total loss: $\mathcal{L}_{\text{total}} \gets \mathcal{L}_\text{MPJPE}$.
\ENDIF
\STATE Backward pass on $\mathcal{L}_{\text{total}}$ and update $\theta$.
\ENDFOR
\ENDFOR
\end{algorithmic}
\end{algorithm}

\subsection{Training Processes and Objectives}
The whole learning process of our method is outlined in Algorithm~\ref{alg}. The MPJPE loss function~\cite{pavllo20193d} $\mathcal{L}_\text{MPJPE}$ is utilized to optimize 3D HPE task:
\begin{equation}\label{loss}
    \mathcal{L}_\text{MPJPE} = \frac{1}{j}\sum_{i=1}^{j}\|\hat{y}_i - y_i\|_2,
\end{equation}
where $j$ denotes the number of human joints. Therefore, the overall loss function during the learning window can be formulated as:
\begin{equation}
    \mathcal{L}_\text{total}= \mathcal{L}_\text{MPJPE}+\mathcal{L}_\text{AWC}.
\end{equation}
\section{Experiments}
\subsection{Experimental Setup}
\noindent{\bf Dataset.}
MM-Fi \cite{yang2024mm} is the first multi-modal non-intrusive 4D human dataset, including four wireless sensing modalities: RGB-D, LiDAR, mmWave radar, and WiFi. This dataset comprises 1,080 video clips, totaling 320k synchronized frames, performed by 40 volunteers engaged in 14 daily activities and 13 rehabilitation exercises. We conduct the experiments on three scenarios outlined in~\cite{yang2024mm} based on activity categories, where Protocol 1 includes 14 daily activities, Protocol 2 includes 13 rehabilitation exercises, and Protocol 3 involves all activities. The S1 random split strategy in~\cite{yang2024mm} is adopted in our experiments.

\noindent{\bf Evaluation Metrics.}
In our experiments, we follow previous works \cite{pavllo20193d} using the following evaluation protocols: Mean Per Joint Position Error (MPJPE) and Procrustes Analysis MPJPE (PA-MPJPE) in millimeters. And MPJPE is formulated as follows:
\begin{equation}
    \text{MPJPE} = \frac{1}{j}\sum_{i=1}^{j}\|\hat{y}_i - y_i\|_2,
\end{equation}
where $j$ is the number of joints.

\noindent{\bf Compared Methods.} 
We compare our method with a joint-training baseline, MM-Fi~\cite{yang2024mm}, X-Fi~\cite{chen2025xfi}, HPE-Li~\cite{d2024hpe}, and several modality balancing methods, including G-Blending \cite{wang2020makes}, OGM-GE \cite{peng2022balanced}, AGM \cite{li2023boosting} and Modality-level resample \cite{wei2024enhancing}, across various fusion strategies: concatenation, MLP, and self-attention.

\noindent{\bf Implementation Details.}
We implement our methods based on Pytorch on two NVIDIA RTX 3090 GPUs. Following \cite{yang2024mm}, we utilize VideoPose3D \cite{pavllo20193d} as the backbone for RGB modality, Point Transformer \cite{zhao2021point} for LiDAR and mmWave, and MetaFi++ \cite{zhou2023metafi++} for WiFi. For the multi-modal learning, we train the model from scratch with randomly initialized parameters in an end-to-end manner, without loading any pre-trained parameters. We employ the Adam optimizer with a momentum of 0.9, a weight decay of 1e-4 and a batch size of 192. The training epoch is 50. The initial learning rate is 1e-3, which is multiplied by 0.1 every 30 epochs.

\begin{table*}[!t]
    \centering
    \caption{Computational overhead breakdown across fusion strategies and modalities. 
    ``Pose Est.'' denotes the total inference time for 3D pose estimation over all modality combinations; 
    ``Correlation'' is the time to compute pose prediction correlation with ground truth; 
    ``Score Calc.'' refers to modality contribution scoring. 
    ``Overhead (\%)'' denotes the score calculation time as a percentage of training time. All times are in milliseconds (ms).}
    \label{tab:shapley}
    \begin{tabular}{lccccccccc}
    \toprule
    Fusion & \#Modalities & \#Params & Forward & Backward & Pose Est. & Correlation & Score Calc. & Overhead (\%) \\
    \midrule
    \multirow{3}{*}{Concat.} 
        & 2 & 15.57 & 104 & 110 & 0.36 & 0.45 & 0.06 & 0.41 \\
        & 3 & 17.77 & 130 & 130 & 0.67 & 0.95 & 0.19 & 0.70 \\
        & 4 & 44.08 & 291 & 445 & 1.33 & 1.61 & 0.43 & 0.46 \\
    \midrule
    \multirow{3}{*}{MLP} 
        & 2 & 16.02 & 109 & 113 & 0.68 & 0.47 & 0.06 & 0.55 \\
        & 3 & 18.33 & 127 & 136 & 1.36 & 0.91 & 0.18 & 0.93 \\
        & 4 & 44.77 & 295 & 450 & 2.91 & 1.69 & 0.45 & 0.68 \\
    \midrule
    \multirow{3}{*}{Attention} 
        & 2 & 18.72 & 110 & 118 & 7.46 & 0.48 & 0.06 & 3.51 \\
        & 3 & 20.90 & 138 & 141 & 13.95 & 0.89 & 0.18 & 5.38 \\
        & 4 & 47.20 & 289 & 460 & 35.81 & 1.76 & 0.48 & 5.08 \\
    \bottomrule
    \end{tabular}
    \vspace{-3mm}
\end{table*}

\subsection{Results and Analysis}
\noindent\textbf{Comparison results of balanced multi-modal learning.} As shown in Table~\ref{table1}, our proposed method outperforms both naive joint training and existing modality balancing techniques, demonstrating its effectiveness in addressing modality imbalance. Specifically, it reduces the MPJPE by approximately 2\,mm and the PA-MPJPE by about 0.5\,mm compared to naive joint training. Moreover, it surpasses other balancing methods by roughly 5\,mm in MPJPE and 2\,mm in PA-MPJPE. As discussed earlier, these competing approaches often overlook the inherent limitations in the information capacity of different modalities. As illustrated in Figure~\ref{vis}, qualitative results from our method closely align with the ground truth across a variety of complex actions, further validating its robustness. Additional results comparing different fusion strategies are provided in the supplementary materials.

\begin{figure}[!t]
    \centering
    \includegraphics[width=\linewidth]{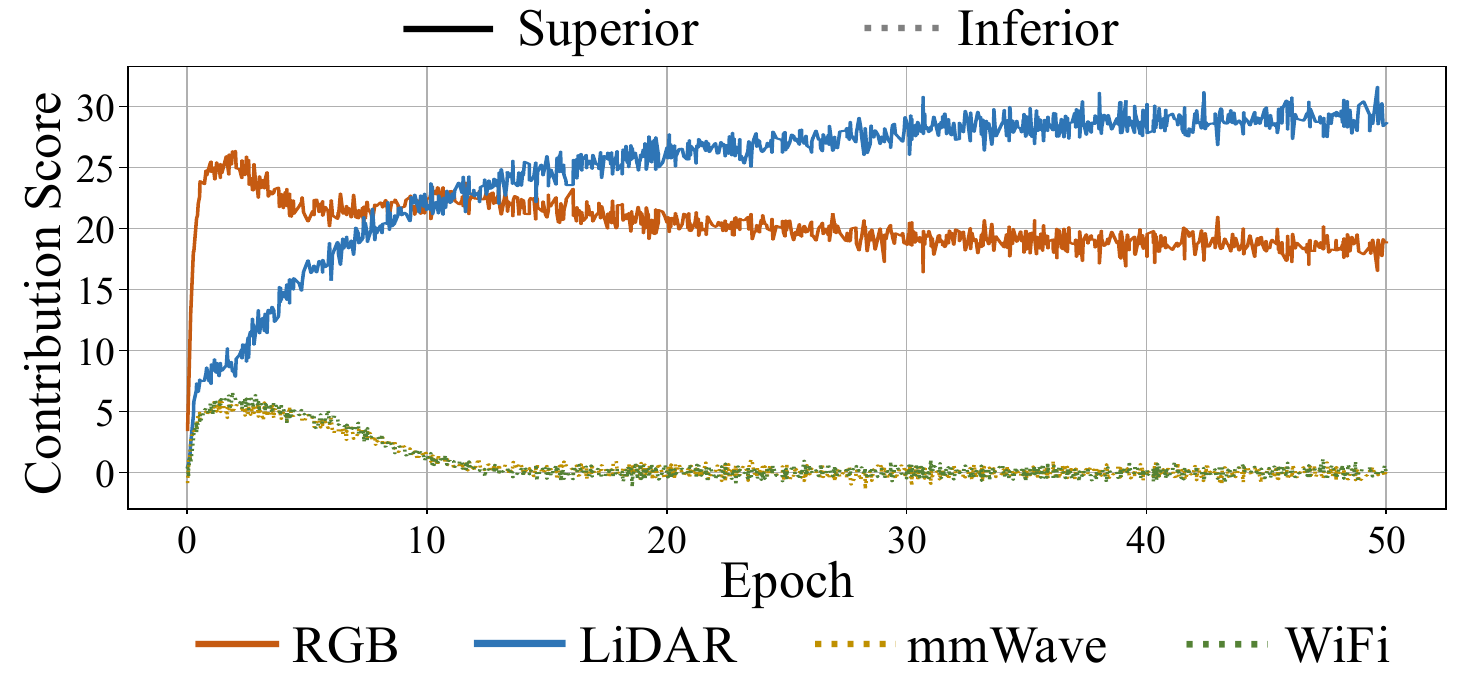}
    \caption{Visualization of contribution scores calculated by our Shapley value-based contribution algorithm using attention-based fusion strategy. }
    \label{score}
    \vspace{-4mm}
\end{figure}

\begin{figure*}[!t]
    \centering
    \includegraphics[width=\linewidth]{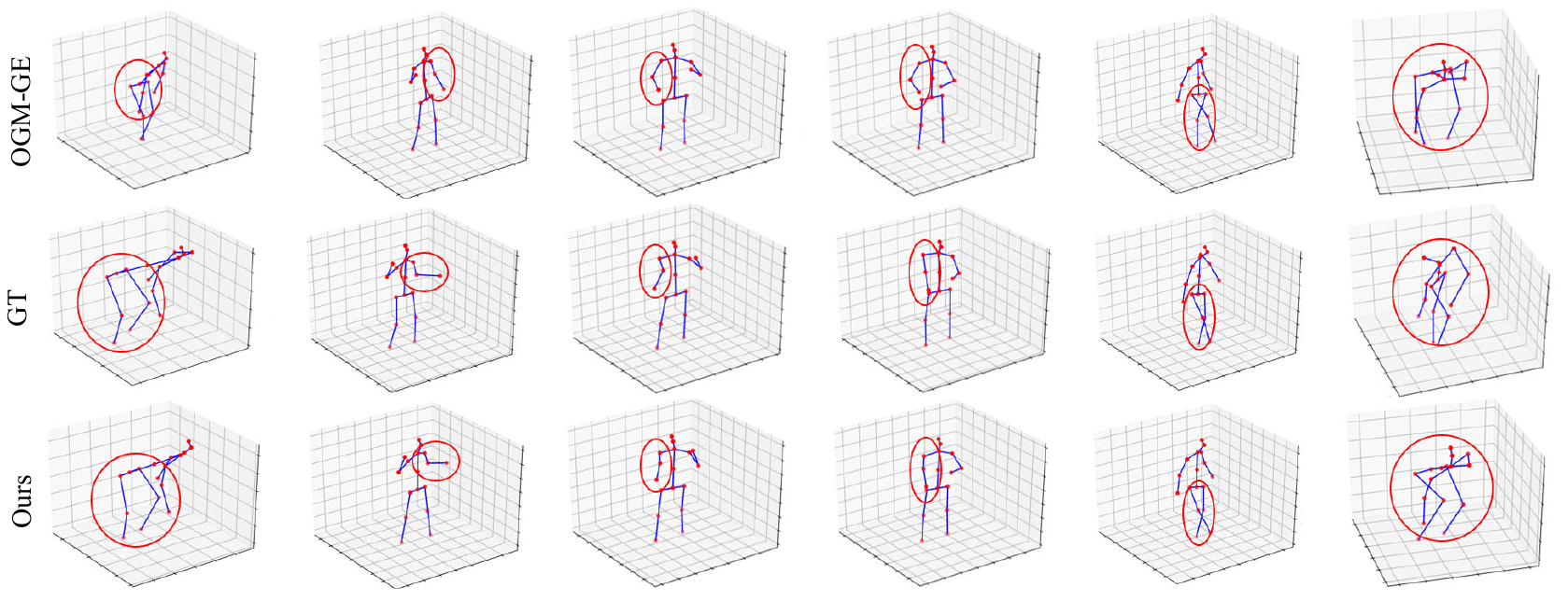}
    \caption{Visual comparisons of 3D human pose estimation between OGM-GE and our method on MM-Fi. Red circles indicate joints where our method achieves superior results.}
    \label{vis}
    \vspace{-3mm}
\end{figure*}

We also conduct experiments to evaluate the contribution scores computed via Shapley values and Pearson correlation. Figure~\ref{score} presents the per-modality contribution scores under Protocol 1. We observe two key phenomena. First, RGB and LiDAR consistently receive higher contribution scores than mmWave and WiFi, indicating their greater utility in the multimodal fusion process. Second, as training progresses, the contribution of inferior modalities gradually diminishes, which suggests that they provide insignificant information for the fusion. In other words, their learning is progressively suppressed during the training. These findings confirm that our scoring mechanism accurately reflects the relative informativeness of each modality, particularly in distinguishing between superior modalities ($\mathcal{M}_{\mathcal{S}}$) and inferior ones ($\mathcal{M}_{\mathcal{I}}$). Further experimental analysis is available in the supplementary material.

\noindent\textbf{Time complexity analysis of Shapley Value.} To evaluate the computational burden introduced by our Shapley value-based modality contribution scoring, we decompose the total runtime into training (forward and backward passes) and post-hoc analysis phases (pose estimation, correlation computation, and score calculation). As defined in Table~\ref{tab:shapley}, the \textit{Overhead (\%)} is computed as the ratio of the total analysis time (Pose Est. + Correlation + Score Calc.) to the training time (Forward + Backward), expressed as a percentage. The results show that the overhead remains remarkably low across all fusion strategies and modality configurations. For instance, under the Concat and MLP fusion schemes, the overhead ranges from only 0.41\% to 0.93\%, even when using up to four modalities. Although the Attention-based fusion incurs higher pose estimation costs due to its iterative cross-modal fusion and reasoning, the total overhead still stays below 5.4\% (e.g., 3.51\% for 2 modalities, 5.38\% for 3, and 5.08\% for 4). While Shapley value computation inherently grows more demanding with an increasing number of modalities, the training cost also rises significantly in such cases. As a result, the relative overhead, defined as the fraction of analysis time over total training time, remains well-contained, demonstrating that our scoring module scales gracefully and imposes only modest computational burden even in multi-modal settings.

\begin{table}[!t]
    \centering
    \caption{Uni-modal and multi-modal performance on Protocol 1.}
    \begin{tabular}{llcc}
    \toprule
    & Modality & MPJPE & PA-MPJPE \\
    \midrule
    \multirow{4}{*}{Uni-modal}
    & RGB & 63.61 & 35.75 \\
    & LiDAR & 66.95 & 45.70 \\
    & mmWave & 102.89 & 52.21 \\
    & WiFi & 166.92 & 97.39 \\
    \midrule
    \multirow{6}{*}{Multi-modal}
    & L+M & 66.85 & 45.11 \\
    & L+W & 68.08 & 46.07 \\
    & M+W & 97.81 & 52.16 \\
    & R+L & \textbf{52.93} & \textbf{34.96} \\
    & R+M & 63.04 & 35.74 \\
    & R+W & 65.68 & 35.89 \\
    & R+L+M+W & \underline{53.87} & \underline{35.09} \\
    \bottomrule
    \end{tabular}
    \label{unimodal}
    \vspace{-3mm}
\end{table}

% \noindent\textbf{Analysis of information capacity.} As illustrated in Table~\ref{unimodal}, RGB and LiDAR consistently outperform the others in both evaluation metrics within the uni-modal setting, indicating their superior information capacity. Notably, RGB achieves low PA-MPJPE close to the all-modality fusion (35.09mm), suggesting that RGB provides precise pose information despite lacking 3D spatial data. Conversely, LiDAR offers rich spatial information but exhibits less precise pose estimation. When combining $\mathcal{M}_{\mathcal{S}}$ (RGB or LiDAR) and $\mathcal{M}_{\mathcal{I}}$ (mmWave or WiFi), the latter provide only marginal performance improvements compared to the fusion of RGB and LiDAR, suggesting that they offer limited information for the task, even when combining with WiFi, the performance decreases compared to uni-modal performance (63.61mm for RGB and *** for RGB+WiFi).

\noindent\textbf{Analysis of modality fusion.}
As shown in Table~\ref{unimodal}, RGB and LiDAR exhibit substantially superior performance compared to mmWave and WiFi in the uni-modal setting, confirming their higher intrinsic information capacity for 3D pose estimation. Notably, RGB alone achieves a PA-MPJPE of 35.75\,mm, which is remarkably close to that of the full four-modality fusion at 35.09\,mm. This demonstrates that RGB can deliver precise pose estimates despite lacking explicit 3D geometric information. Moreover, the combination of LiDAR and RGB substantially improves performance, where LiDAR contributes rich spatial structure even though it yields less accurate joint localization. When fusing superior modalities (RGB and LiDAR) with inferior ones (mmWave and WiFi), the latter contribute only marginal or even detrimental improvements. Crucially, the fusion of all four modalities (R+L+M+W) results in an MPJPE of 53.87\,mm, which is worse than the RGB+LiDAR (R+L) combination that achieves 52.93\,mm. This performance drop provides clear evidence of modality competition. This observation underscores that simply incorporating more modalities does not guarantee better results. Instead, effective multi-modal learning requires explicit mechanisms to balance contributions and suppress interference from modalities.

\begin{table}[t]
    \centering
    \caption{Ablation study on the sensitivity of AWC loss hyperparameters $\alpha_{\mathcal{S}}$ and $\alpha_{\mathcal{I}}$ under Protocol 1.}
    \begin{tabular}{cccc}
    \toprule
    $\alpha_{\mathcal{S}}$ & $\alpha_{\mathcal{I}}$ & MPJPE & PA-MPJPE \\
    \midrule
    \multicolumn{2}{c}{Concat.} & 53.87 & 35.09 \\
    \midrule
    0 & 10k & 52.92 (0.95$\downarrow$) & 34.94 (0.15$\downarrow$) \\
    10k & 0 & 52.09 (1.78$\downarrow$) & 34.81 (0.28$\downarrow$) \\
    10k & 10k & 51.88 (1.99$\downarrow$) & 34.84 (0.25$\downarrow$) \\
    10k & 20k & 51.60 (2.27$\downarrow$) & 34.77 (0.32$\downarrow$) \\
    20k & 0 & 51.99 (1.88$\downarrow$) & 34.77 (0.32$\downarrow$) \\
    20k & 10k & 51.16 (\textbf{2.71}$\downarrow$) & 34.46 (\textbf{0.63}$\downarrow$) \\
    20k & 20k & 51.69 (2.18$\downarrow$) & 34.84 (0.25$\downarrow$) \\
    30k & 20k & 51.34 (2.53$\downarrow$)& 34.56 (0.53$\downarrow$) \\
    \bottomrule
    \end{tabular}
    \label{hyper:alpha}
    \vspace{-3mm}
\end{table}

\begin{figure}
    \centering
    \includegraphics[width=\linewidth]{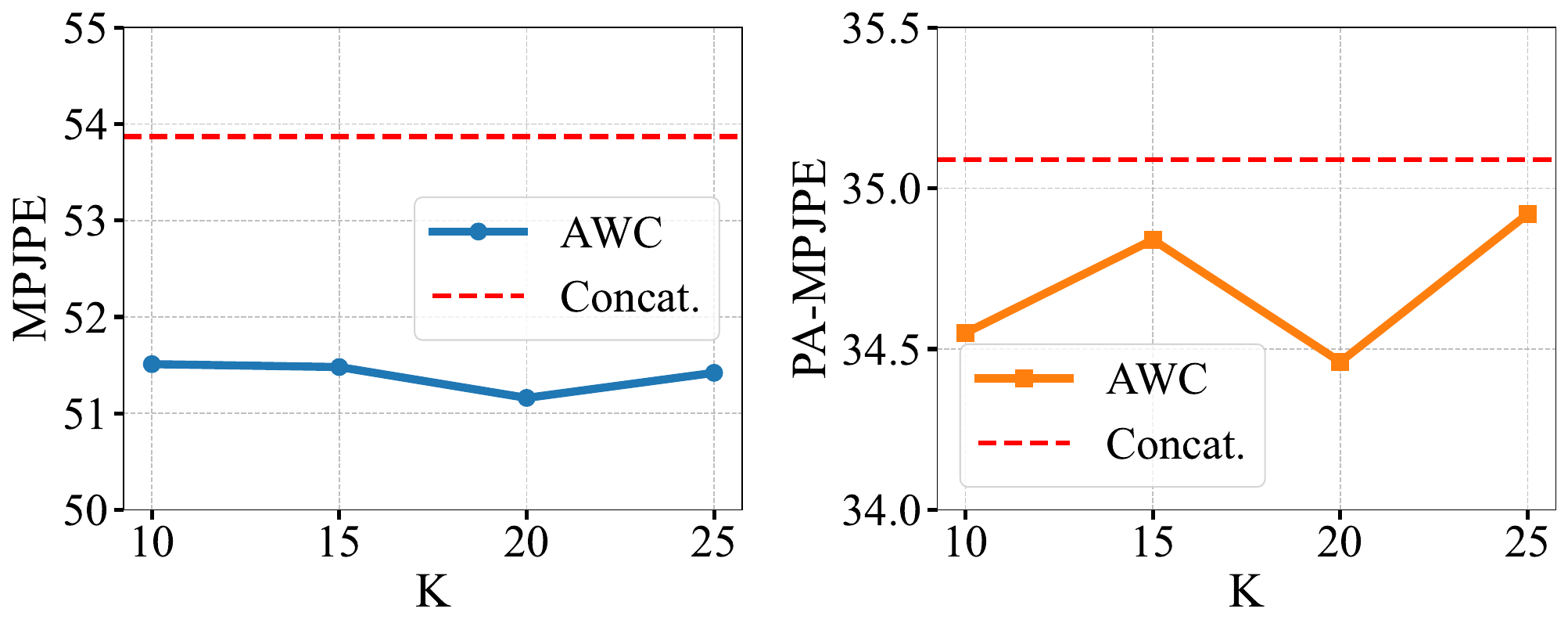}
    \caption{The performance of different $K$ values on Protocol 1.}
    \label{fig:k}
    \vspace{-3mm}
\end{figure}

\noindent\textbf{Hyperparameters sensitivity analysis.} The hyperparameters $\alpha_{\mathcal{S}}$ and $\alpha_{\mathcal{I}}$ in the AWC loss control the regularization strength for dominant and inferior modalities, respectively. As shown in Table~\ref{hyper:alpha}, our method consistently outperforms the \emph{Concat.} baseline across various settings. The best performance is achieved with $\alpha_{\mathcal{S}} = 20k$ and $\alpha_{\mathcal{I}} = 10k$, yielding a 2.71 mm improvement in MPJPE and a 0.63 mm gain in PA-MPJPE. Notably, disabling regularization on inferior modalities ($\alpha_{\mathcal{I}} = 0$) leads to slightly worse results, indicating that constraining both modality groups is essential to enhance representation learning and suppress noise in weaker modalities.
We also analyze the sensitivity of our method to the learning window length $K\in\{10,15,20,25\}$, which controls the regularization epoch. As shown in Figure~\ref{fig:k}, performance peaks at $K=20$, achieving the lowest MPJPE (51.16 \,mm) and PA-MPJPE (34.46 \,mm), indicating that a moderate regularization epoch balances modality learning.

\section{Conclusion}
In this paper, we presented a novel balanced multi-modal learning framework for 3D human pose estimation that effectively addressed modality imbalance. Our approach leveraged a Shapley value–based contribution assessment to dynamically quantify the utility of each modality during training, enabling adaptive modulation of their learning dynamics without auxiliary components or architectural modifications. To further promote balanced optimization, we introduced a weight constraint loss that regularized parameter updates in proportion to their importance, thereby slowing down dominant modalities while safeguarding underrepresented ones from overfitting to noisy signals. Extensive experiments on the MM-Fi dataset validated the superiority and effectiveness of our proposed approach.
\section{Acknowledgement}
This work is partly supported by the Funds for the National Natural Science Foundation of China under Grant 62572072, Beijing Natural Science Foundation under Grant L243027 and NSFC under Grant U24B20176.
{
    \small
    \bibliographystyle{ieeenat_fullname}
    \bibliography{main}

@String(CVPR= {IEEE Conf. Comput. Vis. Pattern Recog.})

@String(ICPR = {Int. Conf. Pattern Recog.})

@String(TOG= {ACM Trans. Graph.})

@String(AAAI = {AAAI})

@String(CVPR  = {CVPR})

@String(ICPR  = {ICPR})

@String(TOG   = {ACM TOG})

@article{yang2024mm,
  title={Mm-fi: Multi-modal non-intrusive 4d human dataset for versatile wireless sensing},
  author={Yang, Jianfei and Huang, He and Zhou, Yunjiao and Chen, Xinyan and Xu, Yuecong and Yuan, Shenghai and Zou, Han and Lu, Chris Xiaoxuan and Xie, Lihua},
  journal={Advances in Neural Information Processing Systems},
  volume={36},
  year={2024}
}

@article{mehta2017vnect,
  title={Vnect: Real-time 3d human pose estimation with a single rgb camera},
  author={Mehta, Dushyant and Sridhar, Srinath and Sotnychenko, Oleksandr and Rhodin, Helge and Shafiei, Mohammad and Seidel, Hans-Peter and Xu, Weipeng and Casas, Dan and Theobalt, Christian},
  journal={Acm transactions on graphics (tog)},
  volume={36},
  number={4},
  pages={1--14},
  year={2017},
  publisher={ACM New York, NY, USA}
}

@article{garcia2019human,
  title={Human 3D pose estimation with a tilting camera for social mobile robot interaction},
  author={Garcia-Salguero, Mercedes and Gonzalez-Jimenez, Javier and Moreno, Francisco-Angel},
  journal={Sensors},
  volume={19},
  number={22},
  pages={4943},
  year={2019},
  publisher={MDPI}
}

@inproceedings{zimmermann20183d,
  title={3d human pose estimation in rgbd images for robotic task learning},
  author={Zimmermann, Christian and Welschehold, Tim and Dornhege, Christian and Burgard, Wolfram and Brox, Thomas},
  booktitle={2018 IEEE International Conference on Robotics and Automation (ICRA)},
  pages={1986--1992},
  year={2018},
  organization={IEEE}
}

@inproceedings{pavllo20193d,
  title={3d human pose estimation in video with temporal convolutions and semi-supervised training},
  author={Pavllo, Dario and Feichtenhofer, Christoph and Grangier, David and Auli, Michael},
  booktitle={Proceedings of the IEEE/CVF conference on computer vision and pattern recognition},
  pages={7753--7762},
  year={2019}
}

@inproceedings{zhao2021point,
  title={Point transformer},
  author={Zhao, Hengshuang and Jiang, Li and Jia, Jiaya and Torr, Philip HS and Koltun, Vladlen},
  booktitle={Proceedings of the IEEE/CVF international conference on computer vision},
  pages={16259--16268},
  year={2021}
}

@article{zhou2023metafi++,
  title={Metafi++: Wifi-enabled transformer-based human pose estimation for metaverse avatar simulation},
  author={Zhou, Yunjiao and Huang, He and Yuan, Shenghai and Zou, Han and Xie, Lihua and Yang, Jianfei},
  journal={IEEE Internet of Things Journal},
  volume={10},
  number={16},
  pages={14128--14136},
  year={2023},
  publisher={IEEE}
}

@inproceedings{peng2022balanced,
  title={Balanced multimodal learning via on-the-fly gradient modulation},
  author={Peng, Xiaokang and Wei, Yake and Deng, Andong and Wang, Dong and Hu, Di},
  booktitle={Proceedings of the IEEE/CVF conference on computer vision and pattern recognition},
  pages={8238--8247},
  year={2022}
}

@inproceedings{hu2022shape,
  title     = {SHAPE: An Unified Approach to Evaluate the Contribution and Cooperation of Individual Modalities},
  author    = {Hu, Pengbo and Li, Xingyu and Zhou, Yi},
  booktitle = {Proceedings of the Thirty-First International Joint Conference on
               Artificial Intelligence, {IJCAI-22}},
  publisher = {International Joint Conferences on Artificial Intelligence Organization},
  editor    = {Lud De Raedt},
  pages     = {3064--3070},
  year      = {2022},
  month     = {7},
  note      = {Main Track},
  doi       = {10.24963/ijcai.2022/425},
  url       = {https://doi.org/10.24963/ijcai.2022/425},
}

@inproceedings{zheng2022multi,
  title={Multi-modal 3d human pose estimation with 2d weak supervision in autonomous driving},
  author={Zheng, Jingxiao and Shi, Xinwei and Gorban, Alexander and Mao, Junhua and Song, Yang and Qi, Charles R and Liu, Ting and Chari, Visesh and Cornman, Andre and Zhou, Yin and others},
  booktitle={Proceedings of the IEEE/CVF Conference on Computer Vision and Pattern Recognition},
  pages={4478--4487},
  year={2022}
}

@inproceedings{cong2022stcrowd,
  title={Stcrowd: A multimodal dataset for pedestrian perception in crowded scenes},
  author={Cong, Peishan and Zhu, Xinge and Qiao, Feng and Ren, Yiming and Peng, Xidong and Hou, Yuenan and Xu, Lan and Yang, Ruigang and Manocha, Dinesh and Ma, Yuexin},
  booktitle={Proceedings of the IEEE/CVF Conference on Computer Vision and Pattern Recognition},
  pages={19608--19617},
  year={2022}
}

@article{geng2022densepose,
  title={Densepose from wifi},
  author={Geng, Jiaqi and Huang, Dong and De la Torre, Fernando},
  journal={arXiv preprint arXiv:2301.00250},
  year={2022}
}

@inproceedings{arshad2017wi,
  title={Wi-chase: A WiFi based human activity recognition system for sensorless environments},
  author={Arshad, Sheheryar and Feng, Chunhai and Liu, Yonghe and Hu, Yupeng and Yu, Ruiyun and Zhou, Siwang and Li, Heng},
  booktitle={2017 IEEE 18th International Symposium on A World of Wireless, Mobile and Multimedia Networks (WoWMoM)},
  pages={1--6},
  year={2017},
  organization={IEEE}
}

@inproceedings{zou2018robust,
  title={Robust WiFi-enabled device-free gesture recognition via unsupervised adversarial domain adaptation},
  author={Zou, Han and Yang, Jianfei and Zhou, Yuxun and Xie, Lihua and Spanos, Costas J},
  booktitle={2018 27th International Conference on Computer Communication and Networks (ICCCN)},
  pages={1--8},
  year={2018},
  organization={IEEE}
}

@inproceedings{wang2020makes,
  title={What makes training multi-modal classification networks hard?},
  author={Wang, Weiyao and Tran, Du and Feiszli, Matt},
  booktitle={Proceedings of the IEEE/CVF conference on computer vision and pattern recognition},
  pages={12695--12705},
  year={2020}
}

@inproceedings{wei2024enhancing,
  title={Enhancing multimodal cooperation via sample-level modality valuation},
  author={Wei, Yake and Feng, Ruoxuan and Wang, Zihe and Hu, Di},
  booktitle={Proceedings of the IEEE/CVF Conference on Computer Vision and Pattern Recognition},
  pages={27338--27347},
  year={2024}
}

@inproceedings{li2023boosting,
  title={Boosting multi-modal model performance with adaptive gradient modulation},
  author={Li, Hong and Li, Xingyu and Hu, Pengbo and Lei, Yinuo and Li, Chunxiao and Zhou, Yi},
  booktitle={Proceedings of the IEEE/CVF International Conference on Computer Vision},
  pages={22214--22224},
  year={2023}
}

@inproceedings{wei2024diagnosing,
  title={Diagnosing and re-learning for balanced multimodal learning},
  author={Wei, Yake and Li, Siwei and Feng, Ruoxuan and Hu, Di},
  booktitle={European Conference on Computer Vision},
  year={2024}
}

@inproceedings{chen2023immfusion,
  title={Immfusion: Robust mmwave-rgb fusion for 3d human body reconstruction in all weather conditions},
  author={Chen, Anjun and Wang, Xiangyu and Shi, Kun and Zhu, Shaohao and Fang, Bin and Chen, Yingfeng and Chen, Jiming and Huo, Yuchi and Ye, Qi},
  booktitle={2023 IEEE International Conference on Robotics and Automation (ICRA)},
  pages={2752--2758},
  year={2023},
  organization={IEEE}
}

@inproceedings{an2022fast,
  title={Fast and scalable human pose estimation using mmwave point cloud},
  author={An, Sizhe and Ogras, Umit Y},
  booktitle={Proceedings of the 59th ACM/IEEE Design Automation Conference},
  pages={889--894},
  year={2022}
}

@inproceedings{he2024video,
  title={Video-Based Human Pose Regression via Decoupled Space-Time Aggregation},
  author={He, Jijie and Yang, Wenwu},
  booktitle={Proceedings of the IEEE/CVF Conference on Computer Vision and Pattern Recognition},
  pages={1022--1031},
  year={2024}
}

@inproceedings{furst2021hperl,
  title={HPERL: 3d human pose estimation from RGB and lidar},
  author={F{\"u}rst, Michael and Gupta, Shriya TP and Schuster, Ren{\'e} and Wasenm{\"u}ller, Oliver and Stricker, Didier},
  booktitle={2020 25th International Conference on Pattern Recognition (ICPR)},
  pages={7321--7327},
  year={2021},
  organization={IEEE}
}

@inproceedings{tekin2016direct,
  title={Direct prediction of 3d body poses from motion compensated sequences},
  author={Tekin, Bugra and Rozantsev, Artem and Lepetit, Vincent and Fua, Pascal},
  booktitle={Proceedings of the IEEE Conference on Computer Vision and Pattern Recognition},
  pages={991--1000},
  year={2016}
}

@inproceedings{pavlakos2017coarse,
  title={Coarse-to-fine volumetric prediction for single-image 3D human pose},
  author={Pavlakos, Georgios and Zhou, Xiaowei and Derpanis, Konstantinos G and Daniilidis, Kostas},
  booktitle={Proceedings of the IEEE conference on computer vision and pattern recognition},
  pages={7025--7034},
  year={2017}
}

@inproceedings{wehrbein2021probabilistic,
  title={Probabilistic monocular 3d human pose estimation with normalizing flows},
  author={Wehrbein, Tom and Rudolph, Marco and Rosenhahn, Bodo and Wandt, Bastian},
  booktitle={Proceedings of the IEEE/CVF international conference on computer vision},
  pages={11199--11208},
  year={2021}
}

@inproceedings{moon2019camera,
  title={Camera distance-aware top-down approach for 3d multi-person pose estimation from a single rgb image},
  author={Moon, Gyeongsik and Chang, Ju Yong and Lee, Kyoung Mu},
  booktitle={Proceedings of the IEEE/CVF international conference on computer vision},
  pages={10133--10142},
  year={2019}
}

@inproceedings{peng2024ktpformer,
  title={KTPFormer: Kinematics and Trajectory Prior Knowledge-Enhanced Transformer for 3D Human Pose Estimation},
  author={Peng, Jihua and Zhou, Yanghong and Mok, PY},
  booktitle={Proceedings of the IEEE/CVF Conference on Computer Vision and Pattern Recognition},
  pages={1123--1132},
  year={2024}
}

@inproceedings{li2024hourglass,
  title={Hourglass Tokenizer for Efficient Transformer-Based 3D Human Pose Estimation},
  author={Li, Wenhao and Liu, Mengyuan and Liu, Hong and Wang, Pichao and Cai, Jialun and Sebe, Nicu},
  booktitle={Proceedings of the IEEE/CVF Conference on Computer Vision and Pattern Recognition},
  pages={604--613},
  year={2024}
}

@inproceedings{xu2024finepose,
  title={FinePOSE: Fine-Grained Prompt-Driven 3D Human Pose Estimation via Diffusion Models},
  author={Xu, Jinglin and Guo, Yijie and Peng, Yuxin},
  booktitle={Proceedings of the IEEE/CVF Conference on Computer Vision and Pattern Recognition},
  pages={561--570},
  year={2024}
}

@inproceedings{cai2019exploiting,
  title={Exploiting spatial-temporal relationships for 3d pose estimation via graph convolutional networks},
  author={Cai, Yujun and Ge, Liuhao and Liu, Jun and Cai, Jianfei and Cham, Tat-Jen and Yuan, Junsong and Thalmann, Nadia Magnenat},
  booktitle={Proceedings of the IEEE/CVF international conference on computer vision},
  pages={2272--2281},
  year={2019}
}

@article{zhu2024probradarm3f,
  title={ProbRadarM3F: mmWave Radar based Human Skeletal Pose Estimation with Probability Map Guided Multi-Format Feature Fusion},
  author={Zhu, Bing and He, Zixin and Xiong, Weiyi and Ding, Guanhua and Liu, Jianan and Huang, Tao and Chen, Wei and Xiang, Wei},
  journal={arXiv preprint arXiv:2405.05164},
  year={2024}
}

@inproceedings{fan2023pmr,
  title={Pmr: Prototypical modal rebalance for multimodal learning},
  author={Fan, Yunfeng and Xu, Wenchao and Wang, Haozhao and Wang, Junxiao and Guo, Song},
  booktitle={Proceedings of the IEEE/CVF Conference on Computer Vision and Pattern Recognition},
  pages={20029--20038},
  year={2023}
}

@book{shapley1953value,
  title={A value for n-person games},
  author={Shapley, Lloyd S and others},
  booktitle = {Annals of Mathematics Studies},
  volume    = {28},
  pages     = {307--317},
  year      = {1953},
  publisher = {Princeton University Press}
}

@inproceedings{wang2015mmss,
  title={Mmss: Multi-modal sharable and specific feature learning for rgb-d object recognition},
  author={Wang, Anran and Cai, Jianfei and Lu, Jiwen and Cham, Tat-Jen},
  booktitle={Proceedings of the IEEE international conference on computer vision},
  pages={1125--1133},
  year={2015}
}

@inproceedings{wu2022characterizing,
  title={Characterizing and overcoming the greedy nature of learning in multi-modal deep neural networks},
  author={Wu, Nan and Jastrzebski, Stanislaw and Cho, Kyunghyun and Geras, Krzysztof J},
  booktitle={International Conference on Machine Learning},
  pages={24043--24055},
  year={2022},
  organization={PMLR}
}

@inproceedings{ReconBoost,
author = {Hua, Cong and Xu, Qianqian and Bao, Shilong and Yang, Zhiyong and Huang, Qingming},
title = {ReconBoost: boosting can achieve modality reconcilement},
year = {2024},
publisher = {JMLR.org},
booktitle = {Proceedings of the 41st International Conference on Machine Learning},
articleno = {789},
numpages = {25},
location = {Vienna, Austria},
series = {ICML'24}
}

@inproceedings{wei2024innocent,
  title={MMPareto: boosting multimodal learning with innocent unimodal assistance},
  author={Wei, Yake and Hu, Di},
  booktitle={International Conference on Machine Learning},
  year={2024}
}

@article{guo2024classifier,
  title={Classifier-guided gradient modulation for enhanced multimodal learning},
  author={Guo, Zirun and Jin, Tao and Chen, Jingyuan and Zhao, Zhou},
  journal={Advances in Neural Information Processing Systems},
  volume={37},
  pages={133328--133344},
  year={2024}
}

@inproceedings{fisher1925theory,
  title={Theory of statistical estimation},
  author={Fisher, Ronald Aylmer},
  booktitle={Mathematical proceedings of the Cambridge philosophical society},
  volume={22},
  pages={700--725},
  year={1925},
  organization={Cambridge University Press}
}

@inproceedings{huang2025adaptive,
  title={Adaptive unimodal regulation for balanced multimodal information acquisition},
  author={Huang, Chengxiang and Wei, Yake and Yang, Zequn and Hu, Di},
  booktitle={Proceedings of the Computer Vision and Pattern Recognition Conference},
  pages={25854--25863},
  year={2025}
}

@inproceedings{achille2018critical,
  title={Critical learning periods in deep networks},
  author={Achille, Alessandro and Rovere, Matteo and Soatto, Stefano},
  booktitle={International conference on learning representations},
  year={2018}
}

@inproceedings{
chen2025xfi,
title={X-Fi: A Modality-Invariant Foundation Model for Multimodal Human Sensing},
author={Xinyan Chen and Jianfei Yang},
booktitle={The Thirteenth International Conference on Learning Representations},
year={2025},
url={https://openreview.net/forum?id=b42wmsdwmB}
}

@inproceedings{d2024hpe,
  title={Hpe-li: Wifi-enabled lightweight dual selective kernel convolution for human pose estimation},
  author={D. Gian, Toan and Dac Lai, Tien and Van Luong, Thien and Wong, Kok-Seng and Nguyen, Van-Dinh},
  booktitle={European Conference on Computer Vision},
  pages={93--111},
  year={2024},
  organization={Springer}
}

@inproceedings{huang2023fuller,
  title={Fuller: Unified multi-modality multi-task 3d perception via multi-level gradient calibration},
  author={Huang, Zhijian and Lin, Sihao and Liu, Guiyu and Luo, Mukun and Ye, Chaoqiang and Xu, Hang and Chang, Xiaojun and Liang, Xiaodan},
  booktitle={Proceedings of the IEEE/CVF International Conference on Computer Vision},
  pages={3502--3511},
  year={2023}
}

@article{qi2025dc,
  title={DC-SAM: In-Context Segment Anything in Images and Videos via Dual Consistency},
  author={Qi, Mengshi and Zhu, Pengfei and Li, Xiangtai and Bi, Xiaoyang and Qi, Lu and Ma, Huadong and Yang, Ming-Hsuan},
  journal={IEEE Transactions on Pattern Analysis and Machine Intelligence},
  year={2025},
  publisher={IEEE}
}

@article{qi2025robust,
  title={Robust disentangled counterfactual learning for physical audiovisual commonsense reasoning},
  author={Qi, Mengshi and Lv, Changsheng and Ma, Huadong},
  journal={IEEE Transactions on Pattern Analysis and Machine Intelligence},
  year={2025},
  publisher={IEEE}
}

@InProceedings{Lv_2026_CVPR,
    author    = {Lv, Changsheng and Fu, Zijian and Qi, Mengshi},
    title     = {Robo-SGG: Exploiting Layout-Oriented Normalization and Restitution Can Improve Robust Scene Graph Generation},
    booktitle = {Proceedings of the IEEE/CVF Conference on Computer Vision and Pattern Recognition (CVPR)},
    month     = {June},
    year      = {2026}
}

@article{qi2025action,
  title={Action quality assessment via hierarchical pose-guided multi-stage contrastive regression},
  author={Qi, Mengshi and Ye, Hao and Peng, Jiaxuan and Ma, Huadong},
  journal={IEEE Transactions on Image Processing},
  year={2025},
  publisher={IEEE}
}

@inproceedings{lv2024sgformer,
  title={Sgformer: Semantic graph transformer for point cloud-based 3d scene graph generation},
  author={Lv, Changsheng and Qi, Mengshi and Li, Xia and Yang, Zhengyuan and Ma, Huadong},
  booktitle={Proceedings of the AAAI Conference on Artificial Intelligence},
  volume={38},
  number={5},
  pages={4035--4043},
  year={2024}
}

@inproceedings{lv2025t2sg,
  title={T2sg: Traffic topology scene graph for topology reasoning in autonomous driving},
  author={Lv, Changsheng and Qi, Mengshi and Liu, Liang and Ma, Huadong},
  booktitle={Proceedings of the Computer Vision and Pattern Recognition Conference},
  pages={17197--17206},
  year={2025}
}

@inproceedings{deng2025global,
  title={Global-local tree search in vlms for 3d indoor scene generation},
  author={Deng, Wei and Qi, Mengshi and Ma, Huadong},
  booktitle={Proceedings of the IEEE/CVF Conference on Computer Vision and Pattern Recognition},
  pages={8975--8984},
  year={2025}
}

@inproceedings{ye2025safedriverag,
  title={Safedriverag: Towards safe autonomous driving with knowledge graph-based retrieval-augmented generation},
  author={Ye, Hao and Qi, Mengshi and Liu, Zhaohong and Liu, Liang and Ma, Huadong},
  booktitle={Proceedings of the 33rd ACM International Conference on Multimedia},
  pages={11170--11178},
  year={2025}
}

@article{wang2025pitn,
  title={PITN: Physics-Informed Temporal Networks for Cuffless Blood Pressure Estimation},
  author={Wang, Rui and Qi, Mengshi and Shao, Yingxia and Zhou, Anfu and Ma, Huadong},
  journal={IEEE Transactions on Mobile Computing},
  year={2025},
  publisher={IEEE}
}

@article{qi2025synergistic,
  title={Synergistic Tensor and Pipeline Parallelism},
  author={Qi, Mengshi and Peng, Jiaxuan and Zhang, Jie and Zhu, Juan and Li, Yong and Ma, Huadong},
  journal={Advances in Neural Information Processing Systems},
  year={2025}
}

@article{qi2021semantics,
  title={Semantics-aware spatial-temporal binaries for cross-modal video retrieval},
  author={Qi, Mengshi and Qin, Jie and Yang, Yi and Wang, Yunhong and Luo, Jiebo},
  journal={IEEE Transactions on Image Processing},
  volume={30},
  pages={2989--3004},
  year={2021},
  publisher={IEEE}
}

@inproceedings{qi2021latent,
  title={Latent memory-augmented graph transformer for visual storytelling},
  author={Qi, Mengshi and Qin, Jie and Huang, Di and Shen, Zhiqiang and Yang, Yi and Luo, Jiebo},
  booktitle={Proceedings of the 29th ACM International Conference on Multimedia},
  pages={4892--4901},
  year={2021}
}

@inproceedings{qi2020imitative,
  title={Imitative non-autoregressive modeling for trajectory forecasting and imputation},
  author={Qi, Mengshi and Qin, Jie and Wu, Yu and Yang, Yi},
  booktitle={Proceedings of the IEEE/CVF Conference on Computer Vision and Pattern Recognition},
  pages={12736--12745},
  year={2020}
}

@article{qi2020stc,
  title={STC-GAN: Spatio-temporally coupled generative adversarial networks for predictive scene parsing},
  author={Qi, Mengshi and Wang, Yunhong and Li, Annan and Luo, Jiebo},
  journal={IEEE Transactions on Image Processing},
  volume={29},
  pages={5420--5430},
  year={2020},
  publisher={IEEE}
}

@article{qi2019sports,
  title={Sports video captioning via attentive motion representation and group relationship modeling},
  author={Qi, Mengshi and Wang, Yunhong and Li, Annan and Luo, Jiebo},
  journal={IEEE Transactions on Circuits and Systems for Video Technology},
  volume={30},
  number={8},
  pages={2617--2633},
  year={2019},
  publisher={IEEE}
}

@article{qi2019stagnet,
  title={StagNet: An attentive semantic RNN for group activity and individual action recognition},
  author={Qi, Mengshi and Wang, Yunhong and Qin, Jie and Li, Annan and Luo, Jiebo and Van Gool, Luc},
  journal={IEEE Transactions on Circuits and Systems for Video Technology},
  volume={30},
  number={2},
  pages={549--565},
  year={2019},
  publisher={IEEE}
}

@inproceedings{qi2019attentive,
  title={Attentive relational networks for mapping images to scene graphs},
  author={Qi, Mengshi and Li, Weijian and Yang, Zhengyuan and Wang, Yunhong and Luo, Jiebo},
  booktitle={Proceedings of the IEEE/CVF conference on computer vision and pattern recognition},
  pages={3957--3966},
  year={2019}
}

@inproceedings{qi2019ke,
  title={Ke-gan: Knowledge embedded generative adversarial networks for semi-supervised scene parsing},
  author={Qi, Mengshi and Wang, Yunhong and Qin, Jie and Li, Annan},
  booktitle={Proceedings of the IEEE/CVF Conference on Computer Vision and Pattern Recognition},
  pages={5237--5246},
  year={2019}
}

@inproceedings{qi2017online,
  title={Online cross-modal scene retrieval by binary representation and semantic graph},
  author={Qi, Mengshi and Wang, Yunhong and Li, Annan},
  booktitle={Proceedings of the 25th ACM international conference on Multimedia},
  pages={744--752},
  year={2017}
}
}


\end{document}